\documentclass[leqno,twocolumn]{article}

\usepackage[left=1.6cm, right=1.6cm, top=1.8cm, bottom=2.1cm]{geometry}

\usepackage{hyperref}

\usepackage{graphicx}
\usepackage{xcolor}
\usepackage{multirow}
\usepackage{booktabs}       
\usepackage{colortbl}
\usepackage{subcaption}
\usepackage[title]{appendix}

\usepackage{xr}
\makeatletter
\newcommand*{\addFileDependency}[1]{
  \typeout{(#1)}
  \@addtofilelist{#1}
  \IfFileExists{#1}{}{\typeout{No file #1.}}
}
\makeatother

\makeatletter
\newcommand{\specificthanks}[1]{\@fnsymbol{#1}}

\begin{document}

\newcommand\relatedversion{}
\renewcommand\relatedversion{\thanks{The full version of the paper can be accessed at \protect\url{https://arxiv.org/abs/1902.09310}}} 


\title{\Large Using NASA Satellite Data Sources and Geometric Deep Learning to Uncover Hidden Patterns in COVID-19 Clinical Severity}
\author{%
  Ignacio Segovia-Dominguez\footnote{NASA Jet Propulsion Lab}
  \and
  Huikyo Lee\textsuperscript{\specificthanks{1}}
  \and
  Zhiwei Zhen\footnote{UT Dallas}
  \and 
  Yuzhou Chen\footnote{Princeton University}
  \and
  Michael Garay\textsuperscript{\specificthanks{1}}
  \and
  Daniel Crichton\textsuperscript{\specificthanks{1}}
  \and
  Rishabh Wagh\textsuperscript{\specificthanks{2}}
  \and
  Yulia R. Gel\textsuperscript{\specificthanks{2}}
}

\date{}

\maketitle







\begin{abstract} \small\baselineskip=9pt 

As multiple adverse events in 2021 illustrated, virtually all aspects of our societal functioning -- from water and food security to energy supply to healthcare -- more than ever depend on the dynamics of environmental factors. Nevertheless, the social dimensions of weather and climate are noticeably less explored by the machine learning community, largely, due to the lack of reliable and easy access to use data.
Here we present a unique not yet broadly available NASA’s satellite dataset on aerosol optical depth (AOD), temperature and relative humidity and discuss the utility of these new data for COVID-19 biosurveillance. In particular, using the geometric deep learning models for semi-supervised classification on a county-level basis over the contiguous United States, we investigate the pressing societal question whether atmospheric variables have considerable impact on COVID-19 clinical severity. 


\end{abstract}


\section{Introduction}






Recent events -- from emergence of new viral pathogens to Texas power crisis to heatwaves of 2021 -- have re-emphasized how vulnerable the security, sustainability, and wellbeing of our society to the human‐induced climate change. Despite the increasing number of initiatives in the machine learning (ML) and data mining (DM) communities to develop more efficient methodology to track climate change and quantify the associated impact on the society~\cite{ICLR2020, ICLR2021, ICML2021}, integration of the state-of-the-art ML and DM methods into climate studies still remains relatively limited, especially, with respect to expansive applications in other disciplines such as social science, bioinformatics, and analysis of cyber-physical systems. 

\begin{figure*}[ht]
\centering
\includegraphics[width=0.95\textwidth]{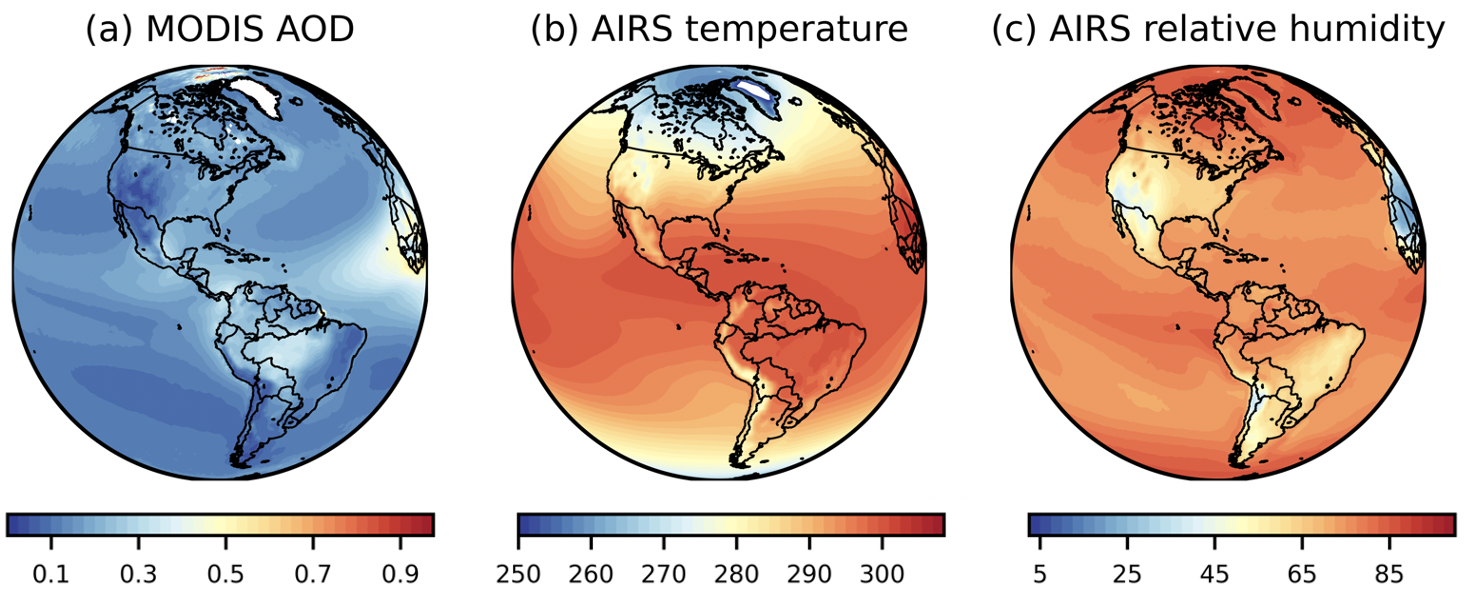} 
\caption{(a) Aerosol optical depth (AOD) in the MODIS averaged for the 19 years between 2001 and 2019. (b) Surface air temperature [K] and (c) relative humidity [\%] in the AIRS averaged for the 17 years between 2003 and 2019.}
\label{Fig:AOD_temp_rh}
\end{figure*}

The ultimate goal of this paper is to make the first step toward bridging this interdisciplinary gap and to bring the unique dataset of NASA’s satellite observations, particularly, on aerosol optical depth (AOD), surface air temperature, and relative humidity to the ML and DM communities. Instruments on NASA’s Earth Observing System (EOS) satellites, especially the MODerate resolution Imaging Spectroradiometer (MODIS)~\cite{levy2013collection} provides high-accuracy measurements of AOD over both land and ocean for the last two decades. These long-term observations of AOD contribute to estimating ground-level PM\textsubscript{2.5} pollution and provide exposure estimates for many epidemiological studies (e.g., ~\cite{Crouse2012, Kloog2013, Franklin2017, Meng2018}). Furthermore, since September 2002, the Atmospheric InfraRed Sounder (AIRS; \cite{Aqua:Aumann:2003}) aboard NASA's Aqua satellite also provides vertical profiles of air temperature and moisture.  Thanks to the broad spatial coverage of AIRS, these observations from AIRS have advanced our understanding of annual cycles in near-surface temperature and moisture all over the globe. Figure \ref{Fig:AOD_temp_rh} shows spatial distributions of AOD, and surface air temperature (temperature afterwards) and relative humidity. These three maps can be drawn by averaging the datasets over time. 

Motivated by the pressing question whether atmospheric conditions exhibit any contribution to COVID-19 dynamics, in this paper we illustrate how the new NASA dataset can be used to evaluate predictive utility and limitations of multiple Recurrent Neural Networks (RNN)-based models and Graph Neural Networks (GNNs) for forecasting COVID-19 clinical severity as a function of atmospheric factors. Indeed, as the recent studies indicate, GNNs and other deep learning (DL) models which are adapted to data in non-Euclidean spaces such as graphs and manifolds and which are often referred to as geometric deep learning (GDL)~\cite{bronstein2017geometric}, can outperform more traditional DL tools in spatio-temporal forecasting tasks. However, systematic comparison of utility and limitations across such GDL models is limited.
In turn, given the irregular lattice structure of the available COVID-19 and other epidemiological data reported at a country level (or county or state levels in the U.S.) and the wide range of uncertainties in the information due to the delayed, incomplete, and noisy official records, GDL using the key spatio-temporal patterns in satellite observations as predictors appears to be one of the most promising forecasting approaches for tracking the hidden mechanisms behind spatio-temporal COVID-19 dynamics.

Making better use of unique NASA’s satellite observations by the ML, DM and, generally, broader scientific community is our top priority. 
Our long-term vision is to provide a one-stop shop for publicly available, easy to use and systematically updated datasets of AOD, temperature, and relative humidity over the entire surface of the Earth  which can be used to address a broad range of ML tasks for social good -- from climate risk mitigation to digital health solutions to fairness in artificial intelligence algorithms for precision farming.

The novelty and contribution of our work can be outlined as follows:
\begin{itemize}
\item To the best of our knowledge, this is the first project addressing potential relationships between atmospheric variables and COVID-19 clinical severity using not only {\it multiple} Geometric DL models but DL, in general. 
Our findings shed a new light on potential environmental risk factors during COVID-19 pandemic, and the new NASA’s satellite benchmark \texttt{NASAdat} lays ground for better understanding sensitivity and robustness of diverse models. 

\item Our \texttt{NASAdat} provides both county-level and state-level information with unique remote sensing features for different regions of the US (i.e., West: CA, South: TX, Northeast: PA) that opens a path for multiple cross-disciplinary applications at the interface of ML, DM and environmental sciences also well beyond COVID-19 surveillance. This includes, for instance, analysis of fairness in healthcare AI algorithms for assessing cancer dynamics in such vulnerable areas as Louisiana Cancer Alley~\cite{mizutani2018backyard, yuan2021assessing}. 

\item Due to urgent and effective actions required to quell the impact of COVID-19 on worldwide, the datasets from NASA’s Distributed Active Archive Centers (DAAC), the data preprocessing technique developed by NASA's Jet Propulsion Laboratory (JPL), and the application of this satellite benchmark datasets shown in this paper provide the guidance for future data collection, the pipeline for remote sensing dataset preprocessing, and model selection.


\end{itemize}

\section{Related Work}

\textbf{Related datasets}. There are few openly available datasets that provides climate data for both research and application purposes. The National Oceanic and Atmospheric Administration (NOAA) \cite{Web:NOAA}, through the National Centers for Environmental Information (NCEI), provides datasets that includes weather variables such as temperature, precipitation, drew point, visibility, etc. However, most of observations in land rely on ground-based stations which limits the resolution on covered areas across U.S., e.g., many counties are far away from land-based stations. Other online services and repositories channel or adapt datasets from previous sources; e.g., National Weather Service \cite{Web:NWeatherS}, Intergovernmental Panel on Climate Change \cite{Web:IPCC-DDC}, WorldClim \cite{Web:WorldClim}, etc. 
In comparison to existing datasets, our daily climatologies of temperature, relative, and humidity provides annual cycles in these three variables for each county with the Federal Information Processing Standard Publication 6-4 (FIPS 6-4) code, as a result, being easier to match with datasets using the same granularity; e.g., COVID-19, Population, Health and Socioeconomic indicators, Mobility, and so on for each county. The original satellite datasets include those three variables at regular Gaussian grids. However, we calculated spatially averaged values in such a way that correspond with county geographical locations.  Finally, long-term AOD observations from a single instrument over the entire CONUS is only available from satellites. Temperature and relative humidity data for the entire globe including those over ocean are another benefit of using satellite observations when running ML models for different spatial domains other than the US. Hence, our proposed dataset, \texttt{NASAdat}, is the first dataset that can be easily used by the broader community to take advantage from NASA's satellite observations. \\

\noindent\textbf{Deep Learning for Spatio-temporal Processes}. One of the most common methods to model multivariate temporal processes are Recurrent Neural Networks (RNN)-based models~\cite{RRN:Yong:2019}. In particular, the family of the Long Short Term Memories (LSTM)~\cite{hochreiter1997long} models demonstrates competitive performance in time series forecasting tasks and capabilities to selectively capture various complex temporal patterns and the Gated Recurrent Units (GRU)~\cite{cho2014learning} as a modification for LSTM for time series forecasting but with simpler computation and implementation. However, RNN-based models are not designed to handle structured data with spatial information, nor naturally extend to non-Euclidean domains. To capture spatial dependency of dynamic structured data and complex spatial and temporal correlations, spatio-temporal dependence-based methods are needed~\cite{li2018diffusion, yu2018spatio, li2019predicting, bai2021a3t, bai2020adaptive, li2019spatio} which have achieved promising results in traffic flow, human motion, and financial time series forecasting.



Recently, Graph Neural Networks (GNNs) and
other GDL tools emerged as a powerful alternative for modeling spatial dependencies in multivariate spatio-temporal processes. In particular,~\cite{li2018diffusion,yu2018spatio,yao2018deep} introduce graph convolution methods into spatio-temporal networks for multivariate time series forecasting which allows us to more accurately capture nonstationary inter- and intra-dependencies among entities~\cite{wu2019graph, bai2020adaptive,cao2020spectral} and to handle data heterogeneity, especially, in nonseparable cases. (By non-separability here, we mean a scenario when spatial dependency varies over time and temporal dependency change with location.) Since
epidemiological data are always reported over the 
irregular polygons of census units, e.g., counties, provinces and states, and also tend to exhibit a highly nontrivial structure of spatio-temporal dependencies due to complexity of socio-environmental and pathogen interactions, 
GDL on manifolds and graphs is a promising new direction for infectious disease mapping. Few previous work use climatological data, from NASA and other sources, along with DL models to explore connections with clinical severity \cite{KDD2021:Segovia-Dominguez:2021,segovia2021tlife}. However, utility of GDL for biosurveillance remains yet largely unexplored and there are no comparative studies on benchmarking GDL, LSTM and other more conventional RNNs for epidemeological applications.


\section{The \texttt{NASAdat} dataset}



The original datasets are publicly available through NASA’s Distributed Active Archive Centers (DAAC) servers. The AIRS3STD product (doi: \url{10.5067/Aqua/AIRS/DATA303}) provides the daily temperature and relative humidity datasets since August 31st, 2002 to the present. 
The daily datasets are provided at regular Gaussian grids with $1^{\circ} \times 1^{\circ}$ resolution. For each day, there is a file containing multiple variables including air temperature and relative humidity at the surface. To prepare our daily climatology data, we downloaded 6209 files for the 6209 days between January 1st, 2003 and December 31st, 2019.    
The Atmosphere Daily Global Product from MODIS on Terra (MOD08\_D3, \url{http://dx.doi.org/10.5067/MODIS/MOD08_M3.006}) contains about 80 variables, including AOD at 550 $nm$ wavelenght, in each file for daily data. AOD from MODIS onboard terra is available from March 2000. We calculated daily AOD climatology using AOD for the 6939 days between January 1st, 2001 and December 31, 2019. To utilize annual cycles in temperature, relative humidity, and AOD in each US county, users of our data need to download only three files from http-urls. 

Appendix~A
provides further details of the generated \texttt{NASAdat} dataset; i.e., data preprocessing, format, DOIs, metadata description, inclusion of county-level FIPS, maintenance plan, uniqueness of \texttt{NASAdat}, and quality control checks.


\section{Biosurveillance for COVID-19: Can Deep Learning Models Quantify Relationships between Atmospheric Factors and COVID-19 Clinical Severity?}  




\begin{figure*}[h!]
  \centering
  \includegraphics[width=1.0\textwidth]{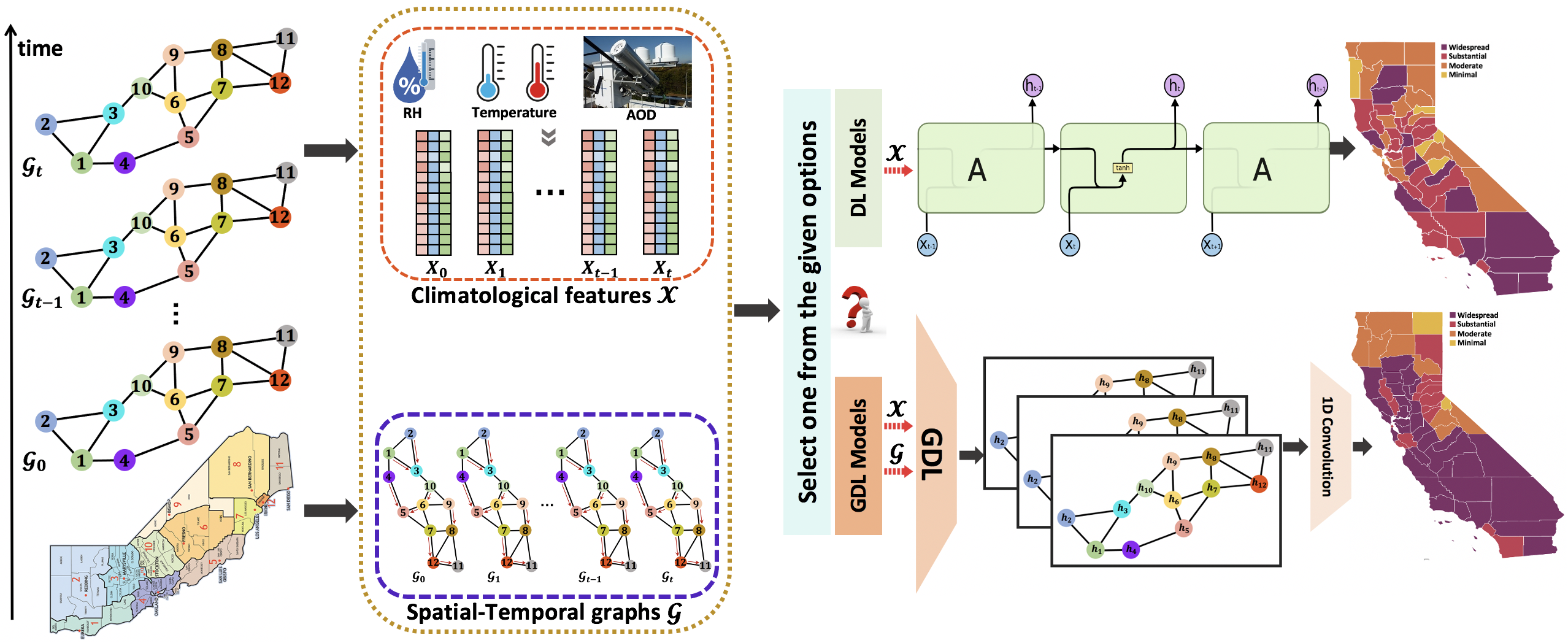}
  \caption{Architecture overview: for multi-step forecasting tasks, two options are provided (i) GDL: the inputs to the GDL models are spatial-temporal graphs with node features (e.g., AOD, Temperature, and RH) and (ii) DL: the input to DL models is node attributes. For GDL, we utilize graph-based convolution neural networks to encode spatial information and temporal information, respectively and then adopt an 1D convolutional layer to compress the output information; for DL, we apply recurrent neural network (with or w/o gated mechanism) to node features for multi-step ahead predictions.}
   \label{Fig:Flowchart}
\end{figure*}

\noindent\textbf{Problem statement} 
Armed with the our proposed benchmark \texttt{NASAdat},
our goal here is to investigate existence of predictive relationships (if any) between atmospheric factors (i.e., satellite observations on AOD, temperature and relative humidity) and COVID-19 clinical severity (measured via COVID-19 hospitalization records).

Formally, $Y_t$ be records on COVID-19 clinical severity and $X_t$ be records on atmospheric variables, $t=1,2, \ldots$. Let a positive integer $h$ be the forecasting horizon. Under the concepts of Granger causality, our objective is to assess how different a conditional distribution of $Y_{t+h}$, given $\{Y_t\}_{t=1}^t$ from a conditional distribution of $Y_{t+h}$, given $\{Y_t, X_t\}_{t=1}^t$, that is, whether time series of atmospheric variables are useful in predicting time series of COVID-19 clinical severity \cite{GrangerCaus:White:2011}.  In practice, this problem is addressed by constructing  the baseline model $M_0$ trained on $\{Y_t\}_{t=1}^t$ and the alternative model  $M_1$ trained on both $\{Y_t\}_{t=1}^t$ and  $\{X_t\}_{t=1}^t$, and then comparing means of the conditional distributions (i.e., RMSE). 

To train Spatio-temporal Graph Convolutional Networks, we represent the connection between adjacency counties as a weighted undirected graph $G = \{V, E, W\}$, where $V$, $E$, and $W$ are the node set, edge set and weight set, respectively. Records of variables $Y_t$ and $X_t$ serve as node features of each node $\nu\in V$. Hence, the remote sensing data (i.e., AOD, temperature, RH) are a part of the node (county) features along with hospitalization numbers. In the experiments, the node feature matrix is fed into the DL/GDL model. That is, GDL propagates and transforms node feature information to learn the node representations and perform final classification/prediction
tasks.



\noindent\textbf{Remark~1} 
Hospitalization and mortality due to COVID-19 are often shown to be closely linked to a prior medical history of lung and other respiratory diseases~\cite{shakil2020covid, gupta2020effect, nunez2021editorial}. Some epidemiological studies have also associated exposure to particulate matter
(PM) air pollution having an aerodynamic diameter smaller than 2.5$\mu m$ (PM2.5) with increased
risk of respiratory diseases~\cite{schraufnagel2019air,liang2009association}.
As such, ambient PM pollution, which has
been estimated with AOD observations, may shed an important light on
assessing and predicting the severity of COVID-19 burden and associated survival rates.
Furthermore, in a post-pandemic world, it will become much more important to evaluate the
implemented strategies for lockdown and vaccine allocation, while accounting for various latent
factors associated with COVID-19 dynamics, in particular, the increasingly more evidenced
impact of polluted air on higher risks of hospitalization due to COVID-19.

\noindent{\bf Remark~2} Note that connectivity among county-geographical locations, e.g. shared borders, provides a natural transmission network to track the disease spread. To account for temporal and spatial dependencies simultaneously, we perform experiments using a wide variety of Recurrent Graph Neural Networks (see Figure~\ref{Fig:Flowchart} for the employed architecture). 


\noindent{\bf Remark~3}. Note that we intentionally do not incorporate any social variables into analysis of COVID-19 clinical severity. First, it is questionable which variables and how impact COVID-19 clinical severity [3]. Second,  our primary focus is to assess conditional predictive utility of atmospheric variables, given that socio-economic, socio-demographic, social-mobility etc factors are fixed. Third, as noted by [1, 2], more polluted areas tend to be populated by economically disadvantaged groups, thereby further increasing unfairness in healthcare outcomes, and it is of critical importance to account for such predisposition in a systematic manner such that the contribution of a single environmental risk factor is analyzed, conditionally on all other factors being fixed. \\

\begin{table*}[h]
\caption{Root Mean Squared Errors (RMSE) for 15-day ahead
forecasts of COVID-19 related hospitalizations, based on DL models in three U.S. states: (a) CA,
(b) PA, and (c) TX, averaged over each
state. Results (RMSE $\pm$ s.d.) are averaged over 10 runs with different seeds; {\bf Bold} indicates the best; while baseline is printed in \textit{italic} font. 
The significance of difference ($p < 0.05$) between {\it Baseline} and {\it Baseline} + \{AOD, Temp, RH\} have been highlighted in \colorbox{yellow}{yellow} via performing with one-sided two-sample $t$-test. The best performance for each state is highlighted in \colorbox{blue!30}{blue}.\label{forecast_result}}
\centering  
\setlength\tabcolsep{2pt}
\resizebox{2.0\columnwidth}{!}{
\begin{tabular}{l|llll|llll|llll}
\toprule
\multirow{2}{*}{\textbf{Model}}& \multicolumn{4}{c|}{\textbf{CA}}& \multicolumn{4}{c|}{\textbf{PA}} & \multicolumn{4}{c}{\textbf{TX}}
\\
\cmidrule(lr){2-5}\cmidrule(lr){6-9}\cmidrule(lr){10-13} 
           & {\it Baseline} & AOD & Temp & RH & {\it Baseline} & AOD & Temp & RH  & {\it Baseline} & AOD & Temp & RH\\
\midrule

LSTM~\cite{hochreiter1997long} & \textit{300.4}{\small$\pm$23.8}& 344.2{\small$\pm$25.0}& 353.9{\small$\pm$5.6}& 352.0{\small$\pm$5.2}& \textit{109.7{\small$\pm$1.8}}& {\cellcolor{yellow!60}93.1{\small$\pm$2.2}}& {\cellcolor{yellow!60}91.0{\small$\pm$1.7}} & {\cellcolor{yellow!60} 98.7{\small$\pm$5.3}}&\textit{73.0{\small$\pm$3.9}} & {\cellcolor{yellow!60}61.6{\small$\pm$5.6}}&{\cellcolor{yellow!60}69.7{\small$\pm$3.2}}
&{\cellcolor{yellow!60}59.3{\small$\pm$4.3}}\\
DCRNN~\cite{li2018diffusion}& \textit{301.6}{\small$\pm$2.2} & {\cellcolor{blue!30}{\bf 221.5{\small$\pm$19.5}}} & 370.3{\small$\pm$6.2}& 351.9{\small$\pm$4.3}& \textit{97.2{\small$\pm$1.6}}& {\cellcolor{yellow!60}94.0{\small$\pm$2.0}}& {\cellcolor{yellow!60}92.2{\small$\pm$2.0}}&97.7{\small$\pm$1.3} & \textit{66.1{\small$\pm$1.4}}& {\bf {\cellcolor{blue!30}49.2{\small$\pm$2.3}}}&79.6{\small$\pm$7.2} & 79.5{\small$\pm$5.2}\\
LRGCN~\cite{li2019predicting} &\textit{295.6}{\small$\pm$26.5} & 344.8{\small$\pm$15.1}&  352.8{\small$\pm$10.2}& 351.5{\small$\pm$5.3}& \textit{109.8{\small$\pm$0.9}}& 115.4{\small$\pm$9.4}&  {\cellcolor{yellow!60}100.5{\small$\pm$5.2}}& {\cellcolor{yellow!60}103.1{\small$\pm$4.0}}& \textit{71.9{\small$\pm$4.7}}& 77.2{\small$\pm$5.7}& {\cellcolor{yellow!60}70.8{\small$\pm$3.6}}
& {\cellcolor{yellow!60}56.9{\small$\pm$5.5}}\\
AT3-GCN~\cite{bai2021a3t} &\textit{257.7}{\small$\pm$7.4} & 370.9{\small$\pm$8.2}& 347.2{\small$\pm$7.5}& 341.7{\small$\pm$4.4}& \textit{88.9{\small$\pm$2.7}}& 95.0{\small$\pm$2.4}& 97.0{\small$\pm$8.9}&{ 99.9{\small$\pm$2.9}} & \textit{66.1{\small$\pm$1.6}}& {\cellcolor{yellow!60}50.8{\small$\pm$1.3}} & {\cellcolor{yellow!60}65.1{\small$\pm$6.8}}& {\cellcolor{yellow!60}57.9{\small$\pm$0.8}}\\
MPNN+LSTM~\cite{panagopoulos2021transfer} & \textit{338.7}{\small$\pm$4.5}& {\cellcolor{yellow!60}302.9{\small$\pm$11.2}}& {\cellcolor{yellow!60}333.4{\small$\pm$7.0}}& 386.9{\small$\pm$14.0}& \textit{109.3{\small$\pm$1.1}}&  {\cellcolor{yellow!60}106.3{\small$\pm$ 1.3}} & 113.5{\small$\pm$4.3}& {\cellcolor{yellow!60}104.3{\small$\pm$2.3}}& \textit{68.4{\small$\pm$1.2}}&69.9{\small$\pm$2.5} & 68.7{\small$\pm$1.8}& 73.1{\small$\pm$1.3}\\
EvolveGCNO~\cite{pareja2019evolvegcn} & \textit{301.5}{\small$\pm$7.2}& 347.7{\small$\pm$1.3}& 339.9{\small$\pm$1.4}& 346.7{\small$\pm$2.1}& \textit{71.5{\small$\pm$17.3}}& 100.2{\small$\pm$3.4}& 89.5{\small$\pm$6.4} & {\cellcolor{blue!30}{\bf53.0{\small$\pm$5.5}}}&\textit{54.6{\small$\pm$3.9}} & 75.8{\small$\pm$0.3}&77.2{\small$\pm$0.2}
& {\cellcolor{yellow!60}51.5{\small$\pm$3.9}}\\
EvolveGCNH~\cite{pareja2019evolvegcn}  & \textit{310.3}{\small$\pm$12.8}& 337.8{\small$\pm$7.4}& 338.9{\small$\pm$10.8}& 310.6{\small$\pm$19.6}& \textit{88.6{\small$\pm$18.7}}& 101.2{\small$\pm$27.7}& 91.1{\small$\pm$29.7} & {\cellcolor{yellow!60}74.1{\small$\pm$21.3}}&\textit{52.7{\small$\pm$5.5}} & 81.8{\small$\pm$8.0}& 84.6{\small$\pm$8.1}
& 57.4{\small$\pm$9.5}\\
GconvLSTM~\cite{seo2018structured} & \textit{310.9{\small$\pm$23.0}} & 330.1{\small$\pm$9.1} &351.6{\small$\pm$10.9} & 342.4{\small$\pm$2.0} & \textit{95.2{\small$\pm$5.7}}	& {\cellcolor{yellow!60}91.2{\small$\pm$2.5}} & {\cellcolor{yellow!60}86.2{\small$\pm$1.6}} & {\cellcolor{yellow!60}94.3{\small$\pm$0.8}} & \textit{56.1{\small$\pm$5.4}} & 62.6{\small$\pm$8.1} & 60.1{\small$\pm$4.6} & 57.8{\small$\pm$5.7}\\
DyGrEncoder~\cite{taheri2019learning}  & \textit{288.5{\small$\pm$25.8}} & 367.8{\small$\pm$25.4} &346.5{\small$\pm$8.4} & 353.3{\small$\pm$9.4} & \textit{107.7{\small$\pm$4.8}}	& {\cellcolor{yellow!60}103.2{\small$\pm$5.8}} & {\cellcolor{yellow!60}95.3{\small$\pm$3.8}} &{\cellcolor{yellow!60}102.5{\small$\pm$7.7}} & \textit{71.5{\small$\pm$4.0}} & {\cellcolor{yellow!60}71.0{\small$\pm$10.5}} & {\cellcolor{yellow!60}65.4{\small$\pm$7.9}} & 70.7{\small$\pm$4.1}\\
\bottomrule
\end{tabular}
}
\end{table*}

\noindent \textbf{Experimental settings}
The methods are trained on a google colab sever with  Intel(R) Xeon(R) CPU @ 2.20GHz, 52 GB RAM, K80,T4 and P100 graphic cards.
We use daily data from February 1st to December 31st of 2020, and split the graph signals into train set, first 80\% of days, and test set, last 20\% of days. Our training step uses 5 lags of daily reported values to produce a 15 days ahead forecasting. We use the AMSGrad optimizer with the same learning rate decay strategy to train all methods with fix learning rate $0.02$ without weight decay. In addition, the dropout rate is 0.5 and epoch number is 500. When training whole architecture, for all methods, we train the model with the same hidden layer dimension ({\it hidden\_dim1} = 128) and the same output dimension ({\it hidden\_dim2} $\in \{55, 60, 251\}$ for CA, PA, and TX respectively). For all methods, we run 10 times in the same partition and report the average accuracy along with standard errors. All experiments are conducted with the following setting: 500 epochs, 1 layer, 128 units, 0.5 as the dropout probability, 0.02 as the learning rate and optimization via AMSGrad. 
Appendix~B 
shows the average running time and standard error of all models on CA, PA, and TX. 
All methods are public which can be found at the \href{https://pytorch-geometric-temporal.readthedocs.io/en/latest/index.html}{PyTorch Geometric Temporal library}~\cite{rozemberczki2021pytorch}. \\

\noindent\textbf{Benchmarking neural network models}
We benchmark two broad classes of neural networks (i) Recurrent Neural Networks (RNNs): 
Long Short-Term Memory (LSTM)~\cite{hochreiter1997long} can forecast univariate time series with LSTM hidden units; (ii) Spatio-Temporal Graph Convolutional Networks: spatio-temporal model with the framework of graph convolutional network (GCN) exploit GCN and temporal convolution to capture dynamic spatial and temporal patterns and correlations; we report performances for eight types of state-of-the-arts methods on our benchmark datasets. Further details in 
Appendix~B.

To the best of our knowledge, this is the first attempt to utilize GDL and, particularly, consensus among \textit{multiple} GDL models for understanding the potential impact of atmospheric variables on COVID-19 hospitalization. (In our earlier research paper \cite{KDDours:UTD-NASA:2021}, we consider a \textit{single} GNN model), and there are not results on addressing COVID-19 hospitalizations on a county level. Furthermore, most recent studies on hospitalization risk prediction employ only (linear) statistical models \cite{HospiRisk::Jehi:2020,Phenotypic:Lasbleiz:2020} and regular convolutional neural networks \cite{CNNFrance:Mohimont:2021}. \\

\begin{figure*}[h!] 
    \centering
    \begin{subfigure}{.425\linewidth}
        \centering
        \includegraphics[width = .75\linewidth]{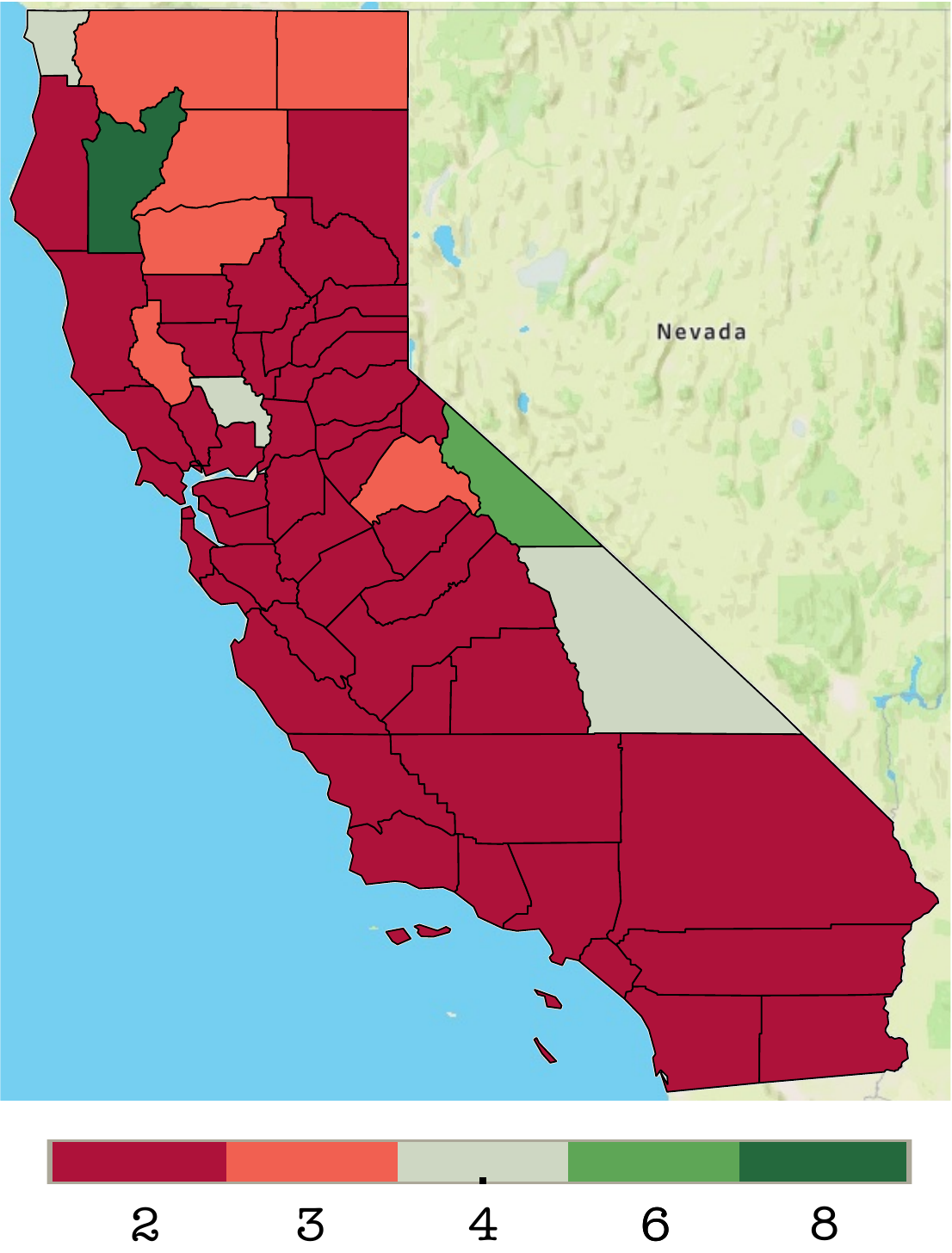}
    \caption{Model Voting with AOD for CA}
    \end{subfigure}    
    \begin{subfigure}{.425\linewidth}
        \centering
        \includegraphics[width = .75\linewidth]{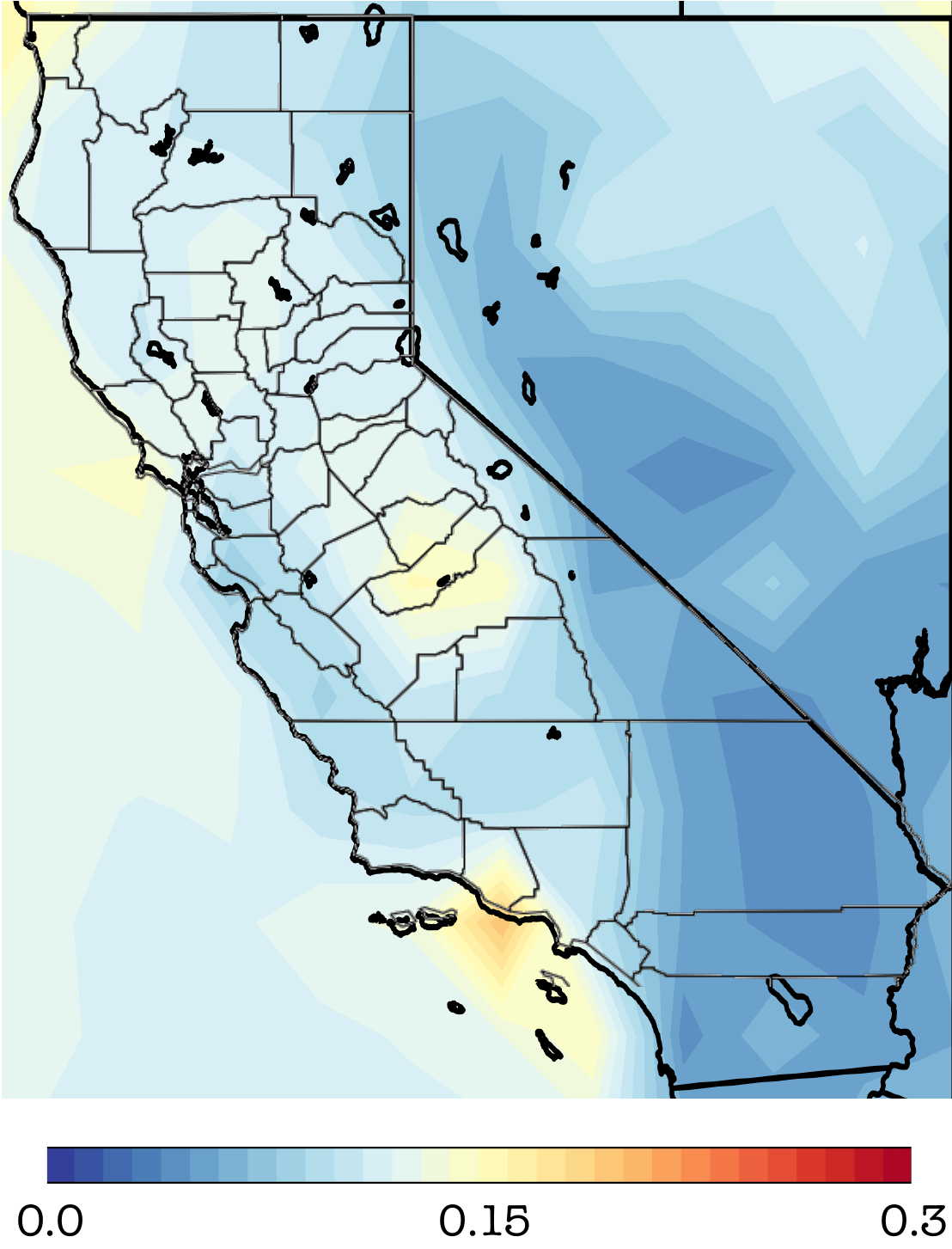}
    \caption{AOD Map for CA}
    \end{subfigure}
    
    \centering
    \begin{subfigure}{.425\linewidth}
        \centering
        \includegraphics[width = .9\linewidth]{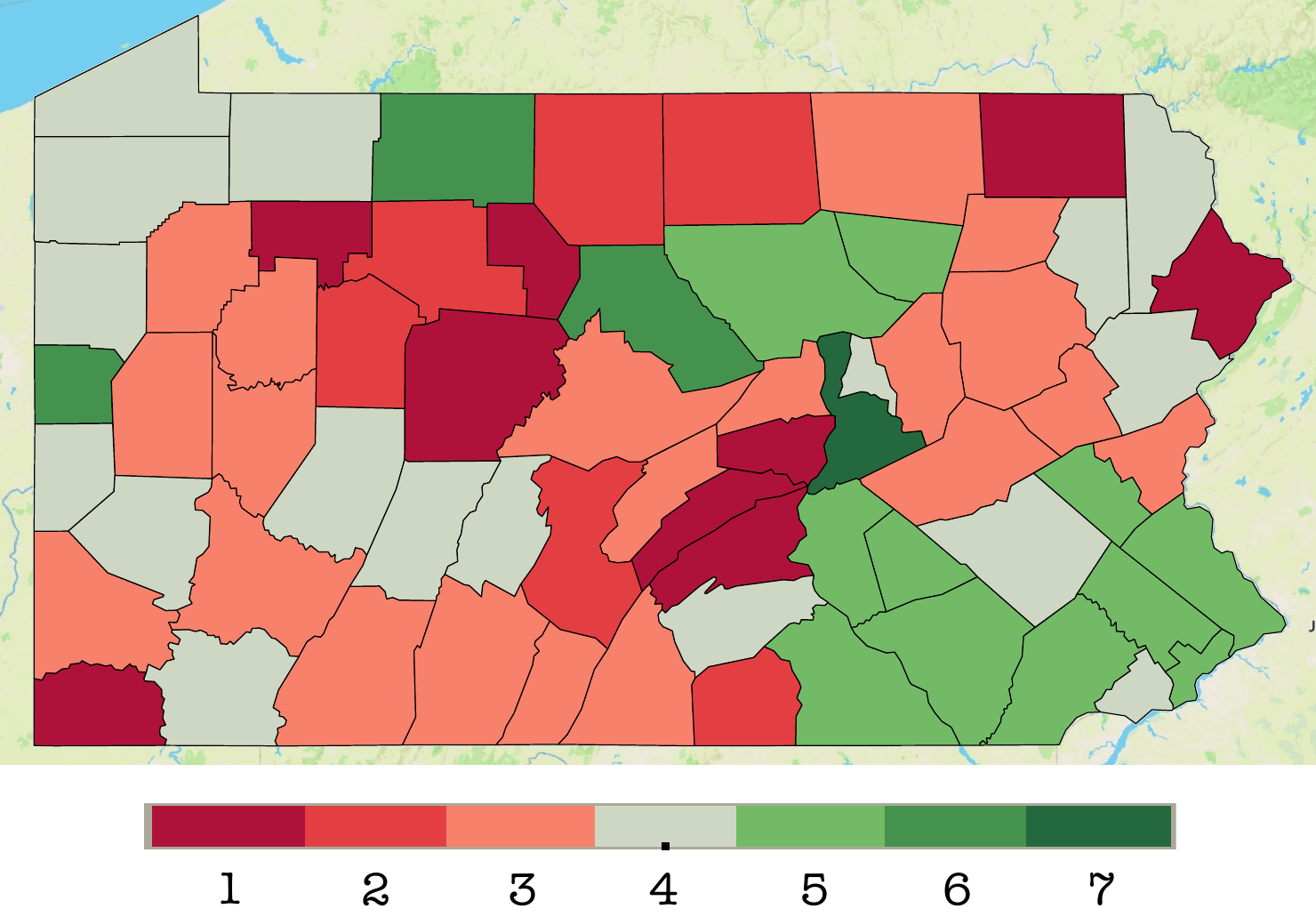}
        \caption{Model Voting with AOD for PA}
    \end{subfigure}
    \begin{subfigure}{.425\linewidth}
        \centering
        \includegraphics[width = .9\linewidth]{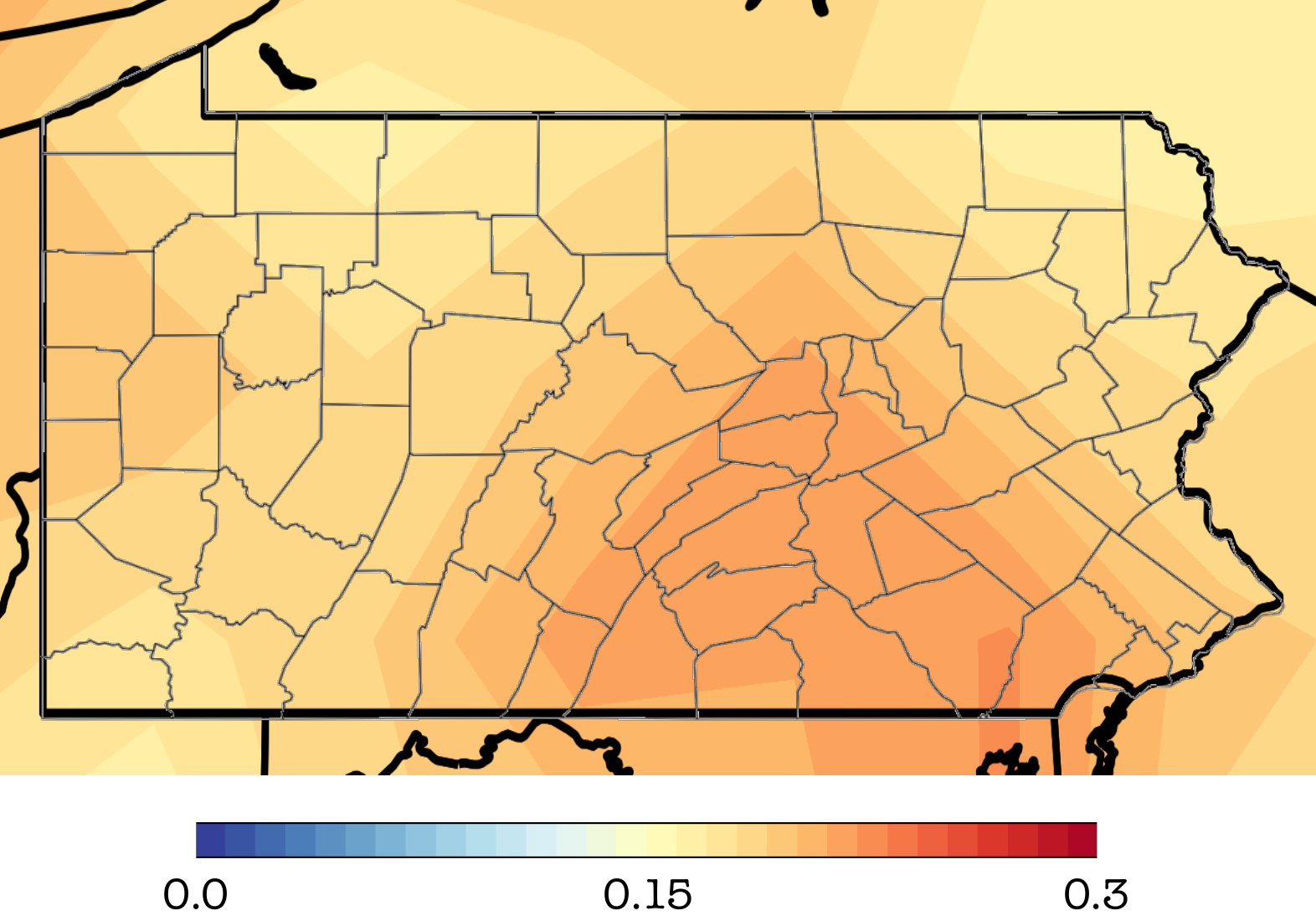}
        \caption{AOD Map for PA}
    \end{subfigure}
    
    \centering
    \begin{subfigure}{.425\linewidth}
        \centering
        \includegraphics[width = .9\linewidth]{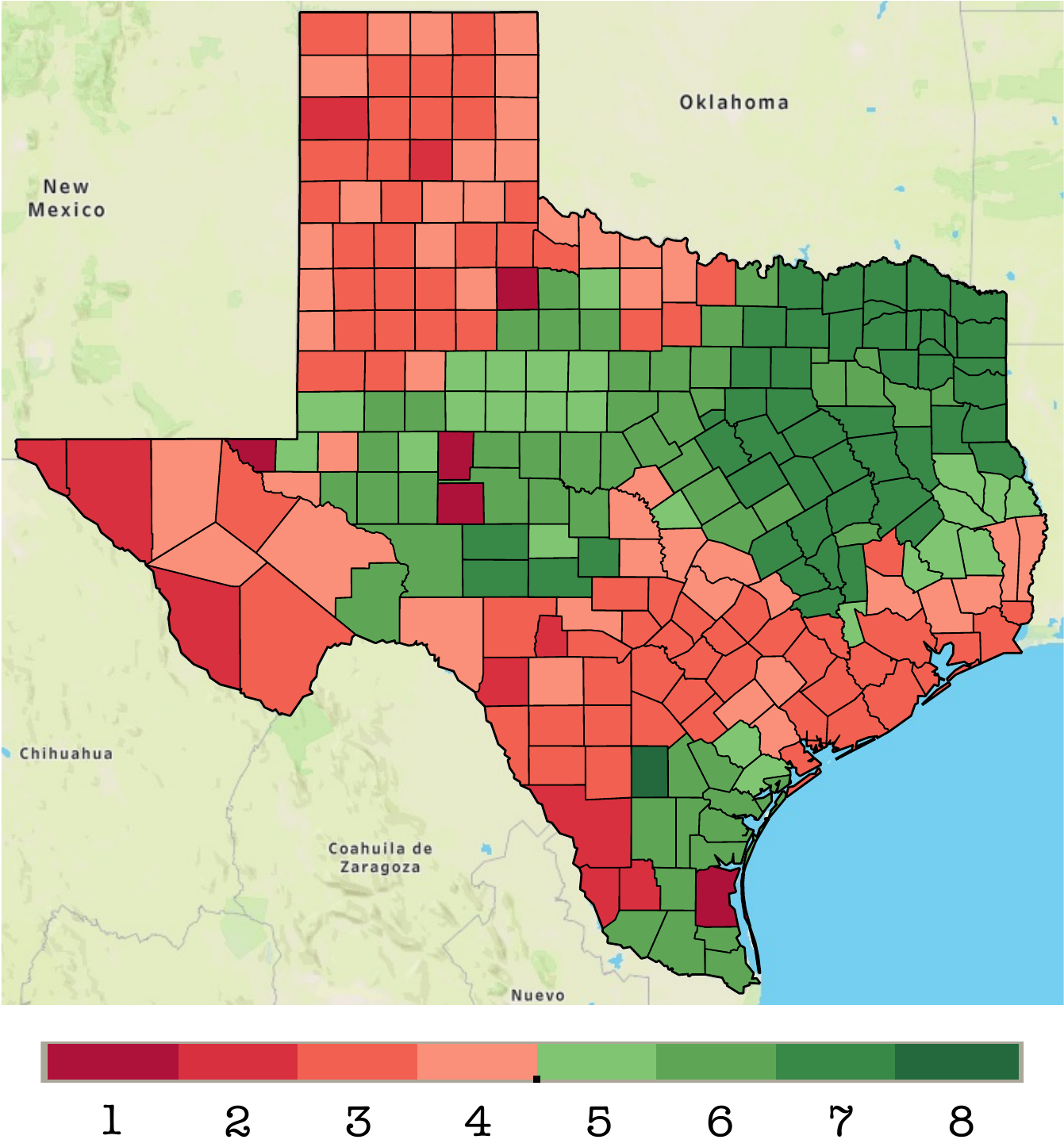}
    \caption{Model Voting with AOD for TX}
    \end{subfigure} 
    \begin{subfigure}{.425\linewidth}
        \centering
        \includegraphics[width = .9\linewidth]{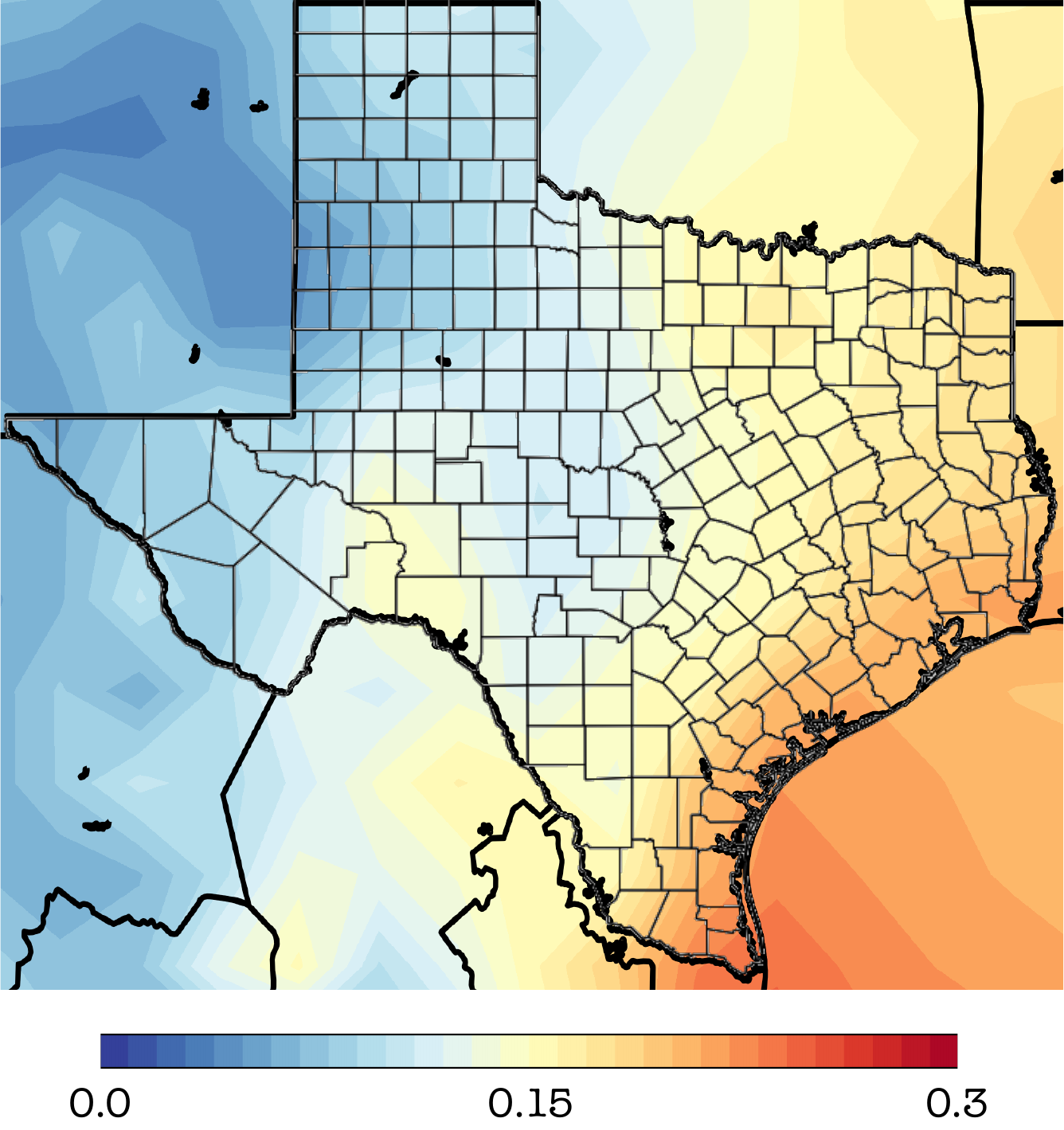}
    \caption{AOD Map for TX}
    \end{subfigure}
    
    \caption{Spatial distribution of the consensus among DL models which find that AOD exhibits predictive utility for COVID-19 clinical severity (a, c, e) and AOD maps (b, d, f) at the county level (i.e., number of model votes agreeing that AOD is a useful predictor for COVID-19 hospitalizations, see 
    Table~3 in Appendix~B).}
    \label{Fig:Counties_CA}
\end{figure*}

\noindent\textbf{Experimental Results} Table~\ref{forecast_result} shows the performance of all benchmarking models on three U.S. states (CA, PA, and TX). For each state, we compare with {\it Baseline} and {\it Baseline} + atmospheric factors (where the {\it Baseline} represents each model in Table~\ref{forecast_result}). Table~\ref{forecast_result} demonstrates the results of DCRNN achieves the best performance with AOD on CA and TX, where DCRNN is based on integrating diffusion convolution into sequence-to-sequence module. Besides, among the models equipped with AOD in the GCN-based class, the competitive results point towards the spectral-based recurrent graph convolution networks, particularly DCRNN and GconvLSTM. In addition, we find that in most cases the {\it Baseline} + atmospheric features outperforms the baseline without  atmospheric features, and the improvements tend to be  statistically significant, especially in PA and TX. These findings indicate that atmospheric factors can help GCN-based model effectively encode the spatial dependencies into a meaningful node representation vector. Overall, our results suggest that emphasizing the importance of modeling the spatial and temporal dependencies outline some avenues for further research on clinical forecasting via dynamics of environmental factors.

Figure~\ref{Fig:Counties_CA} shows spatial distribution of DL model consensus (measured as number of DL model votes for declaring AOD a useful predictor of COVID-19 clinical severity) on a county level for CA, PA, and TX. We find that a very high proportion of the DL models agree that COVID-19 hospitalizations in North Eastern TX, Great Houston and South-Eastern PA are impacted by higher AOD rates. This phenomenon can be attributed to a higher than average number of unhealthy air days in these counties~\cite{ANT}. For instance, Houston experiences 9 times more on-road air pollution than its metro area counterparts~\cite{EDF}, while Philadelphia ranks as one of the 25 worst US metro areas for ozone and year-round particle pollution~\cite{ALA}. Interestingly, a relatively high number of the DL models also agree that AOD impacts COVID-19 hospitalizations in Trinity County of CA. Trinity is the fourth least populated counties in CA with no incorporated cities. We hypothesize that such findings for Trinity might be explained by the potential discrepancies in COVID-19 records or insufficient epidemiological information.  Overall, in contrast to CA which is largely characterized by lower AOD levels and substantially more heterogeneous landscape properties, both PA and TX tend to exhibit higher AOD and higher sensitivity of COVID-19 clinical severity to poor air quality.  


Finally, we also find that LSTM encoder can largely improve the local spatial information propagation (see LSTM, DCRNN, MPNN+LSTM, and DyGrEncoder). In appendix~B
we show further experimental results and analysis on under/over-forecasting of models, and computed time complexity statistics for each DL and GDL models in the three US states: TX, PA, and CA. Additionally, we also illustrate the benefit of including atmospheric factors into GCN-based models percentages of counties in each state, where the DL predictive performance has been improved upon adding (one) atmospheric variable as predictor.
As a future work, 
we plan to feed spectral graph convolution into the LSTM encoder to learn the dynamics of clinical events.

\section{Limitations and Broader Impacts}  

While the presented dataset on AOD, temperature and relative humidity provides multiple opportunities both for the ML community as a spatio-temporal benchmark for GDL, RNNs and other models, and for a broad range of ML for Social Good initiatives, e.g., development of pro-active health risk mitigation strategies, precision agriculture, and resilience quantification, we also foresee a number of limitations. 
First, we currently cannot improve spatial resolution of the presented data and offer records only over USA. 
As AIRS and MODIS on Terra provides longest observation records for the three variables, in the future, it is necessary to fuse datasets at higher spatial resolutions from multiple instruments and provide uncertainty for the fused data. Second, 
since it is known that areas with poorer air quality tend to be often populated by socio-economically disadvantaged groups, 
the results of our study, particularly, in conjunction with the linkages among COVID-19 clinical severity and AOD can dis-proportionally negatively impact such subpopulations in terms of further urban segregation, elevated health and house insurance premiums, and ghost town effects. Appendix~C shows further details about the global impact and applications beyond bio-surveillance.



\section{Discussion}

Despite the considerable impacts of aerosols on the Earth’s radiation budget and air quality, aerosol response to the changing climate and even its mean state in the current climate are poorly represented in climate models according to 
the Intergovernmental Panel on Climate Change (IPCC)~\cite{change2021ipcc}. Hence, NASA's role as provider of significant satellite data for the scientific community is crucial to improve the current modeling tools. Furthermore, this work complements the ongoing and planned missions funded by NASA for better understanding aerosols in Earth’s climate system. Aerosols and their effects on air quality and human health meet the outlined observation strategy of the 2017-2027 Decadal Survey for Earth Science and Applications from Space (ESAS)~\cite{board2018midterm} with the highest priority of “Designated”. 
Nevertheless, NASA's data presented in this paper remain yet largely unavailable to the broader ML community and the ultimate goal of this project is to open a two-way street where both NASA's data are more widely used in various ML tasks and ML initiatives for social good such as biosurveillance, spatio-temporal forecasting, and pattern matching, as well as more state-of-the-art ML algorithms are brought into climate studies.  Finally, we are confident that the proposed benchmark dataset will further enhance the utility of various NASA's instruments to the ML community for present and future problems. \\

\noindent\textbf{Acknowledgement} The project has been supported by NASA grant 20-RRNES20-0021 under the Rapid Response and Novel Research in Earth Science.


\bibliographystyle{plain}
\bibliography{TheReferencesARXIV.bib}

\clearpage

\begin{appendices}










\section{Further Details of the collected \texttt{NASAdat} Dataset}
\label{sec:AppendixNASAdat}

\noindent\textbf{Data Preprocessing and Format} AOD is a measure of the amount of light that atmospheric aerosols scatter and absorb and a monotonic function of air quality related to particulate matter near the ground. We generated daily climatology of AOD using the 19-year observations between January 1st, 2001 and December 31st, 2019 (Figure~1 (a)) 
and used the climatological AOD in the team’s previous studies \cite{KDD2021:Segovia-Dominguez:2021,segovia2021tlife}. To calculate a climatological mean for each day of the year, we average 19 observations between January 1st, 2001 and December 31st, 2019. For example, the climatological AOD on January 1st is an average of the 19 New Year's days from 2003 through 2019. 

We also provide data on daily climatology of surface air temperature and RH from the Atmospheric InfraRed Sounder \cite{Aqua:Aumann:2003} as shown in Figures~1 (b) and~(c). To fully take advantage of its high spatial resolution, we use surface air temperature and relative humidity from AIRS and CrIs in years 2020 and 2021. For example, GDL models can use topological summaries of the Community Long-term Infrared Microwave Coupled Atmospheric Product System (CLIMCAPS) products as input. The underlying hypothesis to be tested over the next three years is that surface air temperature and RH may affect COVID-19 hospitalization and death indirectly.


The collected dataset include a unique identifier for each county and is saved in the netCDF format. \textit{NASAdat} can be accessed via:\\
{\it Temperature}\\
\href{https://doi.org/10.48577/jpl.z31y-2r10}{\color{blue}DOI: 10.48577/jpl.z31y-2r10} \hfill
\href{https://commons.datacite.org/doi.org/10.48577/jpl.z31y-2r10}{\color{blue}Metadata (url) } \\
{\it Relative Humidity} \\
\href{https://doi.org/10.48577/jpl.ws86-1q81}{\color{blue}DOI: 10.48577/jpl.ws86-1q81} \hfill
\href{https://commons.datacite.org/doi.org/10.48577/jpl.ws86-1q81}{\color{blue}Metadata (url)} \\
{\it AOD} \\
\href{https://doi.org/10.48577/jpl.k37v-y751}{\color{blue}DOI: 10.48577/jpl.k37v-y751} \hfill
\href{https://commons.datacite.org/doi.org/10.48577/jpl.k37v-y751}{\color{blue}Metadata (url)}

By including the Federal Information Processing Standard code (FIPS) of each county, now NASA's atmospheric data in \texttt{NASAdat} is easily matched with county level datasets from other public and private entities. \\

\noindent\textbf{Uniqueness} The collected \texttt{NASAdat} dataset is unique in multiple aspects. First, long-term AOD observations from a single instrument over the entire CONUS, such as our \texttt{NASAdat},  is \underline{only} available from satellites. While AOD observations are also available from NASA’s remote sensing Aerosol Robotic Network (AERONET) stations, AERONET coverage is noticeably sparser. In turn, many previous studies which compare AOD observations from MODIS with those from AERONET report reasonable agreement between the two, which also can serve as an additional measure of data quality control.  Second, while NOAA through NCEI provides data on such weather variables as temperature, precipitation, drew point, visibility, etc. Almost all of NOAA’s records rely on ground-based stations. As a result, in contrast to \texttt{NASAdat}, the NOAA data are limited to the resolution on covered areas across U.S., and many counties are far away from land-based stations which further increases uncertainty in applications requiring better resolution, such as biosurveillance.  Third, in comparison to all other  existing data, our daily climatologies of temperature and relative humidity provide annual cycles in these variables for each county with the Federal Information Processing Standard Publication 6-4 (FIPS 6-4) code, thereby making it easier to match \texttt{NASAdat} with various key biosurveillance, socio-economic and socio-demographic information of the best available granularity (i.e., at a county level) such as COVID-19 hospitalizations, cancer rates, and number of houses with solar panels. Fourth, temperature and relative humidity data for the entire globe including those over ocean are another benefit of using satellite observations when running ML models for different spatial domains other than the US. Fifth, the climatology datasets such as \texttt{NASAdat} can be used to study the impacts of the nation’s climate change on various sectors, from digital agriculture to resilience of critical infrastructures to adverse climate events. Moreover, given multiple types of ground truth instances associated with these data, e.g., dust storms and teleconnection patterns, the presented benchmark \texttt{NASAdat} can serve as a test bed for a very broad range of ML tasks such as spatio-temporal forecasting with graph neural networks, transfer learning of climatic scenarios, dynamic clustering, anomaly detection, and multi-resolution pattern matching. \\

\noindent\textbf{Quality of the dataset}
\texttt{NASAdat} undergoes standard data quality control checks under NASA guidelines. The original datasets were generated by averaging quality-controlled observations. As a part of retrieval algorithms, a quality flag is automatically assigned to each retrieved value of temperature, relative humidity, and AOD. The algorithms assign a quality flag of each pixel by comparing the observed values with predefined ranges of valid observations. A quality flag is a kind of automated annotation by a machine that is already considered in the original datasets. As such, we were confident about the quality of our newly generated datasets. Due to low-quality retrievals, there exists a small fraction of missing values in the original datasets. As per the standard statistical practice, these missing values are stripped when calculating a spatial and temporal average for each county. \\
Both MODIS and AIRS missions provide more detailed information on the quality flag.\\
\textit{MODIS}\\ \url{https://atmosphere-imager.gsfc.nasa.gov/sites/default/files/ModAtmo/documents/QA_Plan_C61_Master_2021_09_22.pdf}\\
\textit{AIRS}\\ \url{https://docserver.gesdisc.eosdis.nasa.gov//public/project/AIRS/V7_L2_Quality_Control_and_Error_Estimation.pdf} \\

Both AIRS and MODIS datasets cover the entire globe. The total sizes of 6209 AIRS and and 6939 MODIS files are about 2.5 and 4.2 gigabytes respectively. In our processed data, each file for temperature, relative humidity, or AOD has a size of 95 MB. \\

\noindent\textbf{Maintenance Plan}
Our previous work \cite{RecordAOD:Huikyo:2018} indicates that even 19 years (2001-2019) may not be long enough to define statistically stable AOD climatology. Also, we recognize that continuous updates are the key for these data utilities, especially for biosurveillance and other time sensitive applications. JPL NASA/Caltech will update our datasets 2 times per year and also whenever new versions of the NASA products are released through NASA's Distributed Active Archive Centers (DAACs).

In our maintenance plan we are taking advantage from the fact that these benchmark data are one of the first projects within the most recent broader NASA’s JPL initiative on hosting datasets, such as these and assigning DOIs so there is persistence for papers, and also capturing the raw and any derived results. As such, JPL will continue updating and maintaining these benchmark data under this broader NASA’s initiative, with external access to a hub under the subdomain of jpl.nasa.gov. Our team will keep producing daily temperature, relative humidity, and AOD datasets from AIRS/CrIS and MODIS/VIIRS in a NetCDF format which can serve as input for multiple projects across the ML and atmospheric sciences communities. To take full advantage of the highest spatial resolution, we plan to expand and use level 2 surface air temperature and relative humidity from AIRS and CrIs of next years. With the combination of using NASA front-end servers, NVIDIA DGX clusters at the NASA Center for Climate Simulation, and parallel processing capabilities and elastic scalability of the Advanced Data Analytics Platform (ADAPT) science cloud, we expect to have no issue maintaining our data for years to come as these services will provide us all the resources necessary with no cost to NASAdat end-users.

\section{Further Experimental Results}
\label{AppendixExperiments}

\noindent\textbf{Benchmarking neural network models}
We benchmark two broad classes of neural networks (i) Recurrent Neural Networks (RNNs): 
Long Short-Term Memory (LSTM)~\cite{hochreiter1997long} can forecast univariate time series with LSTM hidden units; (ii) Spatio-Temporal Graph Convolutional Networks: spatio-temporal model with the framework of graph convolutional network (GCN) exploit GCN and temporal convolution to capture dynamic spatial and temporal patterns and correlations; we report performances for eight types of state-of-the-arts methods on our benchmark datasets including (1) Diffusion Convolutional Recurrent Neural Network (DCRNN)~\cite{li2018diffusion}: diffusion convolution recurrent neural network that captures both spatial and temporal dependencies through random walks on graph and encoder-decoder architecture for multi-step forecasting, (2) Long Short-Term Memory R-GCN (LRGCN)~\cite{li2019predicting}: time-evolving neural network which integrates relational GCN (R-GCN) into the LSTM to fully investigate both intra-time and inter-time relations, (3) Attention Temporal Graph Convolutional Network (A3T-GCN)~\cite{bai2021a3t}: an attention temporal GCN that combines GCNs and GRUs with attention mechanism which can capture both spatio-temporal dependencies and global variation trends; (4) Message Passing Neural Networks with LSTM (MPNN+LSTM)~\cite{panagopoulos2021transfer}: a time-series version of message passing neural networks consists of a series of neighborhood aggregation layers to model in detail the dynamics of the spreading process; (5) Evolving Graph Convolutional Networks (EvolveGCNO and EvolveGCNH)~\cite{pareja2019evolvegcn}: evolving graph convolutional network that utilizes the recurrent model to update the trainable parameters of GCN for understanding and forecasting graph structure dynamics; (6) Graph Convolutional Recurrent Network (GconvLSTM)~\cite{seo2018structured}: graph convolutional recurrent network model which replaces convolution by graph convolution to extract the spatial-temporal information; (7) Gated Graph Neural Networks for Dynamic Graphs (DyGrEncoder)~\cite{taheri2019learning}: gated graph neural networks for dynamic graphs which uses a gated graph neural network equipped with standard LSTM for dynamic graph classification. \\

\noindent\textbf{Under/Over Prediction analysis}
Table~\ref{over_prediction} shows fraction of days the DL and GDL models deliver COVID-19 hospitalization forecasts which are higher than the true records (i.e., over-predict). We find that under all considered scenarios the DL and GDL models, with or without atmospheric variables, largely tend to under-forecast COVID-19 related hospitalizations. Such phenomena may be due to inherent model bias and can be addressed by developing ensemble approaches allowing us to more systematically quantify predictive uncertainty. Under-forecasting is particularly prominent for PA and TX. While under- and over-forecasting have its own limitations, it is especially important to account for such under-forecasting phenomena in tasks related to allocation of healthcare resources such ICU beds. \\

\begin{table*}[h]
\caption{Fraction of days (in $\%$) the DL models deliver forecasts higher than the true COVID-19 hospitalization numbers (i.e., over-predict) in three U.S. states: CA, PA, and TX. Forecasting horizon is 15-day ahead. Results (RMSE $\pm$ s.d.) are averaged over 10 runs with different seeds.\label{over_prediction}}
\centering
\setlength\tabcolsep{2pt}
\resizebox{2.0\columnwidth}{!}{
\begin{tabular}{l|cccc|cccc|cccc}
\toprule
\multirow{2}{*}{\textbf{Model}}& \multicolumn{4}{c|}{\textbf{CA}}& \multicolumn{4}{c|}{\textbf{PA}} & \multicolumn{4}{c}{\textbf{TX}}
\\
\cmidrule(lr){2-5}\cmidrule(lr){6-9}\cmidrule(lr){10-13} 
           & {\it Baseline} & AOD & Temp & RH & {\it Baseline} & AOD & Temp & RH  & {\it Baseline} & AOD & Temp & RH\\
\midrule

LSTM~\cite{hochreiter1997long} 
& {23.90}{\small$\pm$3.70 }
& 23.20{\small$\pm$2.82}
& 19.72{\small$\pm$1.40}
& 18.70{\small$\pm$1.37}
& 5.50{\small$\pm$0.63}
& 9.14{\small$\pm$1.63}
& 9.71{\small$\pm$1.24}
& 7.43{\small$\pm$2.84}
& 30.86{\small$\pm$1.23}
& 28.46{\small$\pm$6.86}
&27.51{\small$\pm$2.78}
&17.12{\small$\pm$3.07}\\
DCRNN~\cite{li2018diffusion}
& {19.33}{\small$\pm$0.69 }
& 28.38{\small$\pm$2.85}
& 20.70{\small$\pm$0.70}
& 15.17{\small$\pm$0.42}
& 4.41{\small$\pm$0.33}
& 5.34{\small$\pm$0.74}
& 5.23{\small$\pm$0.55}
& 3.64{\small$\pm$0.70}
& 24.45{\small$\pm$0.49}
& 6.86{\small$\pm$1.54}
& 28.02{\small$\pm$3.95}
& 10.27{\small$\pm$0.44}\\
LRGCN~\cite{li2019predicting} 
& {24.36}{\small$\pm$4.95}
& 19.37{\small$\pm$2.62}
& 20.49{\small$\pm$1.51}
& 19.25{\small$\pm$1.22}
& 3.70{\small$\pm$0.25}
& 1.27{\small$\pm$0.92}
& 2.76{\small$\pm$0.46}
& 3.06{\small$\pm$0.99}
& 30.20{\small$\pm$1.60}
& 1.27{\small$\pm$1.04}
&22.73{\small$\pm$5.28}
&16.48{\small$\pm$3.00}\\
AT3-GCN~\cite{bai2021a3t}
& {28.20}{\small$\pm$0.63}
& 18.97{\small$\pm$0.56}
& 20.23{\small$\pm$0.88}
& 19.81{\small$\pm$0.53}
& 11.05{\small$\pm$0.95}
& 11.69{\small$\pm$0.59}
& 10.51{\small$\pm$2.72}
& 10.28{\small$\pm$0.63}
& 24.65{\small$\pm$0.68}
& 22.92{\small$\pm$2.04}
& 7.82{\small$\pm$1.41}
& 26.07{\small$\pm$1.10}\\
MPNN+LSTM~\cite{panagopoulos2021transfer} 
& {27.19}{\small$\pm$1.27}
& 25.36{\small$\pm$0.88}
& 26.75{\small$\pm$2.45}
& 13.59{\small$\pm$1.84}
& 5.77{\small$\pm$0.12}
& 6.91{\small$\pm$0.42}
& 5.05{\small$\pm$1.37}
& 8.47{\small$\pm$ 0.73}
& 13.82{\small$\pm$0.69}
& 8.48{\small$\pm$3.13}
& 7.81{\small$\pm$2.64}
& 11.46{\small$\pm$1.37}\\
EvolveGCNO~\cite{pareja2019evolvegcn}
& 23.05{\small$\pm$0.69}
& 16.63{\small$\pm$0.55}
& 20.13{\small$\pm$0.83}
& 20.89{\small$\pm$0.47}
& 14.39{\small$\pm$4.90}
& 5.55{\small$\pm$0.32}
& 8.88{\small$\pm$1.32}
& 21.57{\small$\pm$4.38}
& 9.31{\small$\pm$0.76}
& 2.60{\small$\pm$0.14}
& 4.03{\small$\pm$0.61}
& 12.57{\small$\pm$1.17}\\
EvolveGCNH~\cite{pareja2019evolvegcn}  
& {22.22}{\small$\pm$1.20}
& 20.19{\small$\pm$1.89}
& 21.61{\small$\pm$1.77}
& 21.90{\small$\pm$0.72}
& 11.06{\small$\pm$5.09}
& 9.43{\small$\pm$6.24}
& 12.06{\small$\pm$5.38}
& 30.65{{\small$\pm$15.14}} 
& 10.57{\small$\pm$3.60}
& 4.42{\small$\pm$4.75}
& 9.38{\small$\pm$13.63}
& 14.54{\small$\pm$5.54}\\
GconvLSTM~\cite{seo2018structured} 
& {24.64}{\small$\pm$3.29}
& 23.47{\small$\pm$1.28}
& 18.84{\small$\pm$2.65}
& 19.77{\small$\pm$1.30}
& 8.68{\small$\pm$1.31}
& 10.92{\small$\pm$0.93}
& 11.66{\small$\pm$0.91}
& 9.96{\small$\pm$0.77} 
& 20.69{\small$\pm$3.28}
&33.39{\small$\pm$2.77}
&21.45{\small$\pm$4.53}
&17.13{\small$\pm$6.22}\\
DyGrEncoder~\cite{taheri2019learning} 
& {24.14}{\small$\pm$8.06}
& 16.45{\small$\pm$2.63}
& 19.74{\small$\pm$1.44}
& 15.60{\small$\pm$1.32}
& 6.05{\small$\pm$0.70}
& 4.49{\small$\pm$2.56}
& 3.93{\small$\pm$1.88}
& 4.18{\small$\pm$1.63}
& 29.14{\small$\pm$1.46}
& 5.25{\small$\pm$5.25}
& 10.65{\small$\pm$5.86}
& 24.32{\small$\pm$3.05}\\
\bottomrule
\end{tabular}
}
\end{table*}

\noindent\textbf{Time complexity}
Table~\ref{Time_result} shows the average running time and standard error of all models on CA, PA, and TX. \\

\begin{table*}[h]
\caption{Time complexity in seconds, mean and std, for each DL and GDL models in U.S. states: CA, PA, TX.\label{Time_result}}
\centering
\setlength\tabcolsep{2pt}
\resizebox{2.0\columnwidth}{!}{
\begin{tabular}{l|lll|lll|lll}
\toprule
\multirow{2}{*}{\textbf{Model}}& \multicolumn{3}{c|}{\textbf{CA}}& \multicolumn{3}{c|}{\textbf{PA}} & \multicolumn{3}{c}{\textbf{TX}}
\\
\cmidrule(lr){2-4}\cmidrule(lr){5-7}\cmidrule(lr){8-10} 
            & AOD & Temp & RH  & AOD & Temp & RH  & AOD & Temp & RH\\
\midrule
LSTM~\cite{hochreiter1997long}    & 192.40 $\pm$ 0.44 &  197.40 $\pm$ 0.94 & 196.93 $\pm$ 0.57 & 201.66 $\pm$ 0.55 & 206.47 $\pm$ 1.02 & 207.50 $\pm$ 1.13 & 576.75 $\pm$ 2.96  &   585.27 $\pm$ 3.35  & 580.16 $\pm$ 2.70 \\
DCRNN~\cite{li2018diffusion}      & 412.44 $\pm$ 3.80 & 287.70 $\pm$ 1.55  &  286.80 $\pm$ 3.19 & 427.46 $\pm$ 1.16 & 295.72 $\pm$ 1.82 & 296.23 $\pm$ 2.98 & 1250.86 $\pm$ 10.65  & 613.05 $\pm$ 1.16 & 847.26 $\pm$ 7.30   \\
LRGCN~\cite{li2019predicting}     & 707.08 $\pm$ 16.32  &  667.43 $\pm$ 16.46 & 521.15 $\pm$ 11.53 &668.91 $\pm$ 7.99 & 677.56 $\pm$ 11.91 & 577.46 $\pm$ 10.47 & 1286.92 $\pm$ 15.31  & 1295.01 $\pm$ 8.39&  861.55 $\pm$ 40.12   \\
AT3-GCN~\cite{bai2021a3t}         & 704.70 $\pm$ 16.50 &  738.78 $\pm$ 59.67 & 728.41 $\pm$ 2.77 & 753.81 $\pm$ 23.40 & 752.51 $\pm$ 7.51 & 809.73 $\pm$ 129.26 & 3343.85 $\pm$ 31.39  & 3154.69 $\pm$ 141.87 & 2949.90 $\pm$ 31.44    \\
MPNN+LSTM~\cite{panagopoulos2021transfer}  & 713.54 $\pm$ 4.11  & 959.21 $\pm$ 10.67  & 960.71 $\pm$ 11.14  & 949.18 $\pm$ 3.82 & 994.15 $\pm$ 9.35 & 931.53 $\pm$ 24.72 & 3040.30 $\pm$ 35.26 & 3124.62 $\pm$ 74.75 &  3894.68 $\pm$ 51.19   \\
EvolveGCNO~\cite{pareja2019evolvegcn}      & 127.37 $\pm$ 0.41  & 128.40 $\pm$ 2.73  & 131.35 $\pm$ 0.77 & 128.39 $\pm$ 0.44 & 128.10 $\pm$ 0.39 & 131.14 $\pm$ 1.61 & 213.53 $\pm$ 0.33  & 217.54 $\pm$ 3.38 & 220.19 $\pm$ 1.31    \\
EvolveGCNH~\cite{pareja2019evolvegcn}      & 227.10 $\pm$ 2.03  &  224.09 $\pm$ 2.10 & 223.81 $\pm$ 2.14 & 228.85 $\pm$ 2.28 & 228.38 $\pm$ 1.61 & 230.25 $\pm$ 3.49 & 348.20 $\pm$ 4.40  & 344.20 $\pm$ 2.34 &  342.21 $\pm$ 2.50   \\
GconvLSTM~\cite{seo2018structured}         & 627.27 $\pm$ 2.06 &  625.42 $\pm$ 0.62 & 618.33 $\pm$ 0.92 & 640.47 $\pm$ 1.20 & 637.65 $\pm$ 1.06 & 630.42 $\pm$ 1.02  & 1188.53
$\pm$ 1.74 & 1189.27 $\pm$ 5.56  & 1183.04 $\pm$ 1.08  \\
DyGrEncoder~\cite{taheri2019learning}      & 367.77 $\pm$ 25.47   & 353.97 $\pm$ 5.66  & 352.04 $\pm$ 5.29 & 103.16 $\pm$ 5.83  & 95.27 $\pm$ 3.77  & 102.46 $\pm$ 7.73  &  73.03 $\pm$ 10.52  & 65.36 $\pm$ 7.97  & 70.69 $\pm$ 4.10 \\
\bottomrule
\end{tabular}
}
\end{table*}

\noindent\textbf{Percentage of improvements}
To further illustrate the benefits of including the atmospheric factors into GCN-based models for COVID-19 hospitalizations analysis, we also
present percentages of counties in each state, where the DL predictive performance has been improved upon adding (one) atmospheric variable as predictor  (see Table~\ref{Improvement_result}). As Table~\ref{Improvement_result} suggests, compared to Temp and RH, most GCN-based models with AOD tend to deliver noticeably higher numbers of counties where
the DL model performance improvement is recorded. These findings also echo the results in Table~1 and reconfirm the earlier hypothesis that AOD is an important factor for understanding COVID-19 clinical severity. \\

\begin{table*}[h]
\caption{Fraction of counties (in $\%$) in three US states: CA, PA, and TX where the  forecasting performance for COVID-19 hospitalizations has been improved upon adding atmospheric variables from the proposed \texttt{NASAdat} dataset into DL models. Highest fractions for each state are in bold.\label{Improvement_result}}
\centering
\setlength\tabcolsep{8pt}
\resizebox{1.5\columnwidth}{!}{
\begin{tabular}{l|lll|lll|lll}
\toprule
\multirow{2}{*}{\textbf{Model}}& \multicolumn{3}{c|}{\textbf{CA}}& \multicolumn{3}{c|}{\textbf{PA}} & \multicolumn{3}{c}{\textbf{TX}}
\\
\cmidrule(lr){2-4}\cmidrule(lr){5-7}\cmidrule(lr){8-10} 
            & AOD & Temp & RH  & AOD & Temp & RH  & AOD & Temp & RH\\
\midrule
LSTM~\cite{hochreiter1997long}   & 5.45 & 5.45 & 5.45 & 88.33 & 90.00 & 78.33 & 82.47 & 33.86 & 47.41  \\
DCRNN~\cite{li2018diffusion}      & {\bf 92.73}  & 1.82  & 3.64  & 23.33  & 33.33  & 26.67  & {\bf 96.41}  & 10.36  & 4.38  \\
LRGCN~\cite{li2019predicting}     & 9.09  & 5.45  & 5.45  & 6.67  & 58.33  & 65.00  & 65.74  & 31.08  & 75.70  \\
AT3-GCN~\cite{bai2021a3t}         & 7.27  & 5.45  & 5.45  & 1.67  & 1.67  & 1.67  & 71.71  & 73.31  & 59.36  \\
MPNN+LSTM~\cite{panagopoulos2021transfer}  & {\bf 92.73}  & 10.91  & 5.45  & {\bf 98.33}  & 75.00  & 90.00  & 41.43  & 42.23  & 10.36  \\
EvolveGCNO~\cite{pareja2019evolvegcn}      & 7.27  & 7.27  & 5.45  & 6.67  & 11.67  & 95.00  & 3.19  & 3.19  & 86.06  \\
EvolveGCNH~\cite{pareja2019evolvegcn}      & 5.45  & 3.64  & 30.91  & 0.00  & 48.33  & 61.67  & 3.19  & 1.20  & 47.41  \\
GconvLSTM~\cite{seo2018structured}         & 9.09  & 5.45  & 5.45  & 53.33  & 73.33  & 35.00  & 64.54  & 15.54  & 64.94  \\
DyGrEncoder~\cite{taheri2019learning}      & 10.91  & 12.73  & 10.91  & 91.67  & 95.00  & 80.00  & 45.02  & 49.40  & 26.29  \\
\bottomrule
\end{tabular}
}
\end{table*}

\noindent\textbf{Further comments on experimental settings}
We do not explicitly incorporate social variables \textit{by purpose}. First, it is questionable which variables and how impact COVID-19 clinical severity \cite{CovidTrans:Ming:2021}. Second, even if someone smokes (potentially a negative factor), does it mean that smokers in polluted and unpolluted areas have the same COVID-19 prognosis? To address such questions, we focus on assessing conditional predictive utility of atmospheric variables, given that socio-economic, socio-demographic, social-mobility etc factors are fixed. Third, as noted by \cite{PolutionPoverty:Lucile:2021,PollutionPoverty:Persico:2020}, more polluted areas tend to be populated by economically disadvantaged groups, thereby further increasing unfairness in healthcare outcomes, and it is of critical importance to account for such predisposition in a systematic manner such that the contribution of a single environmental risk factor is analyzed, conditionally on all other factors being fixed.

{\bf Why RMSE?} In our experiments we use the RMSE metric rather than $R^2$ since RMSE is the standard metric for validation of predictive models in space-time forecasting \cite{BookTimeSeries:Brockwell:1991}. 
Despite statistical criticism, $R^2$ is still used in epidemiology.  As such, we present a summary of results for $R^2$. While we
find that $R^2$ for actual observations and hospitalization forecasts with/without AOD are generally similar in CA, in TX and PA $R^2$ for GCNs \textit{with} AOD tends to be from 0.05 to 0.25 higher than $R^2$ for the same GCN but \textit{without} AOD, with ranges from 0.6 to 0.88 in PA and from 0.71 to 0.93 in TX. These findings echo our conclusions on contributions of AOD to COVID-19 clinical severity, based on predictive RMSE.

{\bf Why Not Regression Models?} Furthermore, we do not consider simpler models, such as regression, ARIMA and other Box-Jenkins class of models, because such tools focus only on linear relationships between variables and, as a result, cannot capture nonlinear nonseparable spatio-temporal dependencies of COVID-19 dynamics (and many other infectious diseases with high virulence). In turn, our analysis includes a broad range of DL architectures that allow us to address such nonlinear dependencies. Furthermore, the model consensus analysis presented in our paper enables us to address such pressing question as  whether 
a relative risk to be affected by COVID-19
is higher for some areas due to their higher exposure to poor air quality.

\section{Global Impact and Applications Beyond Bio-surveillance}
\label{sec:AppendixImpact}

Let us emphasize that NASA’s satellite observations such as \texttt{NASAdat} have played an important role in the international assessment reports on climate change that provide scientific resources for understanding human-induced climate change and assessing impacts crucial for informing policy. As such, the climatology datasets such as \texttt{NASAdat} can be used to study the impacts of the nation’s climate change on various sectors, from digital agriculture to resilience of critical infrastructures to adverse climate events. Moreover, given multiple types of ground truth instances associated with these data, e.g., dust storms and teleconnection patterns, the presented benchmark \texttt{NASAdat} can serve as a test bed for a very broad range of ML tasks such as spatio-temporal forecasting with graph neural networks, transfer learning of climatic scenarios, dynamic clustering, anomaly detection, and multi-resolution pattern matching. For example, scientists can explore transferability of the developed topological and geometric machine learning approaches to monitoring and forecasting spread and clinical severity of (re)emerging SARS- and MERS-associated viruses and arboviral diseases such as Zika. Alternatively, we can study how transferable is the response of the Dallas power grid network to Texas’ winter storms to other adverse climate events and/or across power grid networks in other US cities. 

Although the current NASAdat benchmark is limited to the CONUS, it is important to take advantage of a global coverage in the observations from polar-orbiting satellites. Prior to application of \texttt{NASAdat} worldwide, it is also important to test the data over the CONUS where ground-based observations of temperature and relative humidity observations are available from NOAA. Not surprisingly, the comparison between the AIRS observations and those from NOAA’s network would show some difference. However, based on the previous studies conducted at JPL, we are confident that there is almost negligible difference in annual cycles of these two variables. 

The biggest advantage of using satellite observations is their wide coverage over the entire globe. We already prepared the same datasets of temperature, relative humidity, and AOD averaged for each country, but have not applied to them to model COVID-19 clinical severity outside the United States yet. Such COVID-19 biosurveillance analysis using GDL and other DL applied to the worldwide dataset will be our future work. Furthermore, we have experimented with multi-resolution pattern matching using topological ML in application to the worldwide dataset \cite{MultiResolutionAOD:Ofori-Boateng:2021}, and we think that such multi-resolution pattern matching will be of interest not only in environmental sciences but in broader problems of image processing and computer vision.

\end{appendices}


\end{document}